%% file: main.tex
\documentclass{article}
\usepackage{ijcai25}

\usepackage{times}
\usepackage{soul}
\usepackage{url}
\usepackage[hidelinks]{hyperref}
\usepackage[utf8]{inputenc}
\usepackage[small]{caption}
\usepackage{graphicx}
\usepackage{amsmath}
\usepackage{amsthm}
\usepackage{booktabs}
\usepackage{algorithm}
\usepackage{algorithmic}
\usepackage{multirow}
\usepackage[table,xcdraw]{xcolor}
\usepackage{tabularx}
\usepackage{amssymb}
\usepackage{makecell}

\usepackage{tikz}
\usepackage{hyperref}
\usetikzlibrary{shapes.geometric, arrows, mindmap}
\usetikzlibrary{positioning}  

\newtheorem{definition}{Definition}

\title{Survey on Strategic Mining in Blockchain: A Reinforcement Learning Approach}

\author{
Jichen Li$^{1,2\dagger}$\and
Lijia Xie$^{1\dagger}$\and
Hanting Huang$^{1,4}$\and
Bo Zhou$^1$\and
Binfeng Song$^{1,4}$\and \\
Wanying Zeng$^{1,3}$\and
Xiaotie Deng$^{1,2*}$\And
Xiao Zhang$^{1,3,5*}$\\
\affiliations
$^1$Zhongguancun Laboratory\\
$^2$Center on Frontiers of Computing
Studies, Computer Science Department, Peking University\\
$^3$LMIB and School of Mathematical Sciences, Beihang University\\
$^4$LMIB and Institute of Artificial Intelligence, Beihang University\\
$^5$Hangzhou International Innovation Institute of Beihang University\\
\emails
\{limo923, xiaotie\}@pku.edu.cn,
\{xielj, zhoubo\}@zgclab.edu.cn,\\
\{hantinghuang, zb2342110, zengzeng, xiao.zh\}@buaa.edu.cn
}

\begin{document}

\maketitle
\footnotetext{$^{\dagger}$Equal Contribution.}
\footnotetext{$^{*}$Corresponding Authors.}
\begin{abstract}
Strategic mining attacks, such as selfish mining, exploit blockchain consensus protocols by deviating from honest behavior to maximize rewards.
Markov Decision Process (MDP) analysis faces scalability challenges in modern digital economics, including blockchain.
To address these limitations, reinforcement learning (RL) provides a scalable alternative, enabling adaptive strategy optimization in complex dynamic environments.

In this survey, we examine RL's role in strategic mining analysis, comparing it to MDP-based approaches. We begin by reviewing foundational MDP models and their limitations, before exploring RL frameworks that can learn near-optimal strategies across various protocols.
Building on this analysis, we compare RL techniques and their effectiveness in deriving security thresholds, such as the minimum attacker power required for profitable attacks. 
Expanding the discussion further, we classify consensus protocols and propose open challenges, such as multi-agent dynamics and real-world validation.

This survey highlights the potential of reinforcement learning (RL) to address the challenges of selfish mining, including protocol design, threat detection, and security analysis, while offering a strategic roadmap for researchers in decentralized systems and AI-driven analytics.
\end{abstract}


\input{Sections/New-1-Intro}
\input{Sections/New-2-MDP_Model}

\input{Sections/New-3-RL}

\input{Sections/New-4-Consensus_Strategy}

\section{Conclusion}
\label{sec:conclusion}
This survey examines strategic mining in blockchain systems, with a focus on reinforcement learning as a tool for optimizing mining strategies.
Traditional Markov Decision Process (MDP) approaches are useful for analyzing behaviors like selfish mining but face scalability challenges.
Reinforcement learning (RL) provides adaptability in complex environments, enabling the identification of optimal strategies and security thresholds. 
This survey reviews previous studies that use MDPs and RL to analyze PoW and PoS consensus, and discusses the potential of these methods for analyzing other vote-based and parallel confirmation blockchains. 
Future research should focus on refining RL algorithms to improve blockchain security and efficiency, ultimately advancing decentralized systems.



\bibliographystyle{named}
\bibliography{ref}

\end{document}

%% file: Sections/New-1-Intro.tex
\section{Introduction}
\label{sec:intro}
In recent years, blockchain technology has been widely applied to solve problems in various domains, developing innovative solutions previously considered impossible.  
It all began with an inventive ledger design to record all transactions generated in a decentralized system called Bitcoin, invented by \cite{nakamoto2008bitcoin}.  
This revolutionary ledger design maintains a sequentially growing list of blocks, each containing several transactions and linked to the preceding block using cryptographic techniques.  
The process of adding new blocks is governed by a consensus mechanism called Proof of Work (PoW), in which participants, called miners, use their computational power to calculate a hash function.
The miner who finds a valid solution for this hash function can have the right to produce a new block and earn rewards for generating it.  
In the design of the Bitcoin protocol, each miner is expected to broadcast the generated block to the network immediately.  
As long as all participants behave honestly, the expected revenue will be proportional to their computational power.  

However, in practice, miners 
are economically rational and profit-seeking.
They may adopt strategic behaviors, implying that honestly broadcasting a block is not necessarily the most rewarding strategy.
This is indeed the case, as evidenced by the study in \cite{eyal2014majority}, which introduced the selfish mining strategy—arguably the most notorious mining attack in blockchain.
In this attack, a miner strategically delays broadcasting the blocks they mine, causing other miners to generate blocks at invalid positions and inducing honest miners to waste their mining power.  
As a result, the strategic miner can earn more (expected) revenue than their fair share. 
Pushing this approach to the extreme, \cite{sapirshtein2016optimal} expanded the action space of selfish mining, modeling it as a Markov Decision Process (MDP), and analyzed the optimal mining strategy for a miner when facing other honest miners.  
A series of works have since been initiated to study mining strategies using this approach, such as \cite{feng2019selfish,grunspan2020selfish,li2021new,ferreira2022optimal}.  

However, with the continuous updates of blockchain consensus and the introduction of new protocols, directly computing a miner's strategy using MDP faces computational difficulties. 
To address this issue,~\cite{hou2019squirrl} proposed a generalizable framework for using reinforcement learning (RL) to analyze blockchain incentive mechanisms.  
Using this approach, researchers only need to model the states and strategies from the miner's perspective in an MDP and then use machine learning methods to learn an approximate optimal strategy.  
This method provides a framework for the detailed analysis of various blockchain protocols and attack patterns, including~\cite{bar2022werlman,bar2023deep,sarenche2024deep}.

In this survey, we provide a comprehensive overview of blockchain strategic mining analysis.  
We first summarize the MDP modeling approaches to analyze miners' strategic mining behavior and review the resulting security thresholds for attackers in different types of consensus protocols.  
Next, we focus on summarizing the existing findings on miners' strategic behavior using RL methods and compare the learning techniques employed as well as the resulting security thresholds.  
Finally, we introduce the MDP modeling paradigms for other consensus protocols in blockchain, such as voting-based and parallel confirmation protocols. 
We also propose several open problems and discuss the potential of using reinforcement learning to analyze these protocols.

\paragraph{The differences between our survey and others.}
Past surveys on strategic mining mainly focus on how to prevent mining attacks.
~\cite{madhushanie2024selfish} focuses solely on the harms of selfish mining attacks and analyzes existing detection and mitigation methods,
while ~\cite{nicolas2020blockchain} focuses on defending against double-spending attacks and selfish mining attacks, proposing various defense strategies.
However, no existing work has systematically surveyed the analytical methods for strategic mining. 
Our paper fills this important gap.

\paragraph{Roadmap.}
In Section~\ref{sec:MDP}, we introduce the MDP modeling method for consensus strategies and the definition and results of the security threshold.  
Then, in Section~\ref{sec:RL}, we present the results of using RL methods to analyze the strategy for mining blocks.  
After that, in Section~\ref{sec:consensus}, we provide an overview of the classification of blockchain consensus mechanisms and how analysis methods are applied within each category. 
Finally, we summarize this survey in Section~\ref{sec:conclusion}.  

%% file: Sections/New-2-MDP_Model.tex
\section{Strategic Mining Analysis via Markov Decision Processes}
\label{sec:MDP}

\begin{table*}[ht]
\centering
\large 
\resizebox{\textwidth}{!}{ 
    \begin{tabular}{| m{4cm}<{\centering}  m{3cm}<{\centering}  m{3.5cm}<{\centering}  m{3cm}<{\centering}  m{4.5cm} |} 
        \hline
        \textbf{Literature} & \textbf{Security Threshold} & \textbf{Method} & \textbf{Blockchain Consensus} & \textbf{Description} \\ \hline
        \cite{eyal2014majority} & 0.25 & Markov Reward Process & Bitcoin PoW & Introduce original selfish mining strategy. \\ \hline
        \cite{sapirshtein2016optimal} & 0.232 & MDP & Bitcoin PoW & Compute the optimal strategy for selfish mining. \\ \hline
        \cite{marmolejo2019competing} & 0.26297 & MDP + Game Theory & Bitcoin PoW & Consider scenario of multiple non-colluding semi-selfish miners. \\ \hline
        \cite{feng2019selfish}  & 0.26 & Two-dimensional MDP & Ethereum PoW  & Propose a two-dimensional MDP to model selfish mining in Ethereum. \\ \hline
        \cite{zur2020efficient}  & 0.2468 & Average Reward Ratio MDP & Ethereum PoW & Propose the Average Reward Ratio (ARR) MDPs, reducing the complexity of computing optimal strategy. \\ \hline
    \end{tabular}
} 
\caption{Studies on security threshold of blockchain consensus using Markov Decision Process}
\label{tab:MDPsec-threshold}
\end{table*}

In this section, we review the MDP modeling approach for consensus strategies, as well as the definition and analysis of security thresholds. 
We begin by examining previous work on the theoretical foundations of MDPs, highlighting key components such as states, actions, transitions, and rewards, which serve as the basis for decision-making in uncertain environments. 
We then explore the application of MDP models to strategic mining, demonstrating how they can optimize mining strategies in consensus protocols. 
Finally, we summarize the security threshold analysis through MDPs, focusing on 
how these models can be utilized to evaluate and to enhance security measures.
This section offers both theoretical insights and practical applications of MDPs in addressing strategic mining problems.
The results for the security threshold of strategic mining are summarized in Table~\ref{tab:MDPsec-threshold}. 

\subsection{Theoretical Foundations of MDP Modeling}

We present the definition of MDP and its fundamental concepts as follows.

\begin{definition}[Markov Decision Process (MDP)]
An MDP is formally defined as a quintuple 
\begin{equation}
    MDP = (S, A, P, R, \gamma),
    \label{eq:MDP_def}
\end{equation}
where:
\begin{itemize}
    \item $S$: Finite or countable state space.
    \item $A$: Finite action-space available from each state.
    \item $P(s' \mid s, a): S \times S \times A \to [0, 1]$, the transition probability matrix, denoting the probability of transitioning from state $s$ to state $s'$ with action $a$.
    \item $R(s, a): S \times A \to \mathbb{R}$, the reward function providing the immediate reward obtained from state $s$ with action $a$.
    \item $\gamma \in [0, 1]$: The discount factor, controlling the importance of future rewards. 
\end{itemize}
\end{definition}

A policy $\pi(a | s)$ is a mapping that defines the probability actions should take in each state.
To calculate the total reward under a given policy $\pi$ in MDP, let $G_t$ be the cumulative reward starting from time $t$ with start state $s_t$, denoted by
\begin{align*}
G_t = \sum_{k=0}^{\infty} \gamma^k \mathbb{E}_{\pi}[R\bigl(s_{t+k}, \pi(s_{t+k})\bigr)],
\end{align*}
where $s_{t+k}$ is the state reached at time step $t+k$. For any state \( s \in S \), the value function \( V(s) \) is the expected cumulative reward calculated according to the current policy \( \pi \). Given the state \( s_t \), the agent selects the action $a_t$ with probability $\pi(a_t | s_t)$. Therefore, the form of the value function is as follows:
\begin{align*}
V^\pi(s) = \mathbb{E}_{\pi}\left[G_t \mid s_t = s\right].
\end{align*}
Furthermore, under a given policy \( \pi \), the expected cumulative reward after performing action \( a \) in state \( s \) is defined as the state-action value function \( Q^\pi(s, a) \), given by the form as
\begin{align*}
Q^\pi(s, a) = \mathbb{E}_{\pi}\left[ R(s_{t}, a_{t}) + \gamma G_{t+1} \mid s_t = s, a_t = a \right].
\end{align*}

The MDP model exhibits the following core properties:
\begin{itemize}
    \item \textit{Markov Property:} MDP has the \textit{memoryless} property, indicating that the transition from the current state depends only on the present state and not on past historical states. Therefore, for states $s_t$ and $s_{t+1}$, it holds that:
    \begin{align*}
        P(s_{t+1} \mid s_t, a_t) = P(s_{t+1} \mid s_0, a_0,  \dots, s_{t}, a_t).
    \end{align*}
    
    \item \textit{Bellman Recursive:} For a given MDP with fixed policy \( \pi \), $V^\pi(s)$ and $Q^\pi(s, a)$ satisfies the Bellman equation:
    \begin{align*}
    V^\pi(s) = \sum_a \pi(a|s) \sum_{s'} P(s'|s,a) \left[ R(s,a) + \gamma V^\pi(s') \right],
    \end{align*} 
    {\small\begin{align*}
        & Q^\pi(s,a) 
        = \sum_{s'}  P(s'|s,a)\left[ R(s,a) + \gamma \sum_{a'} \pi(a'|s') Q^\pi(s',a')\right].
    \end{align*}}
    
    \item \textit{Stationary Distribution:} If the state space is finite and the transition matrix is ergodic, then there exists a stationary distribution $\phi(s)$ under deterministic policy $a = \pi(s)$ satisfied
    \begin{align*}
    \phi(s') = \sum_{s} \phi(s) P(s' \mid s, \pi(s)),
    \end{align*}
    where $\phi(s)$ describes the long-run probability of each state and is an eigenvector of the transition matrix.
\end{itemize}

\subsection{MDP Model for Strategic Mining}
The seminal work on selfish mining attack~\cite{eyal2014majority} considers a system with two types of miners: selfish miner and honest miner. 
Let $\alpha$ denote the fraction of mining power possessed by the selfish miner. 
The block generation process is modeled as a random process, where a new block is generated in each time slot. 
In this scenario, selfish miners maintain a private chain after mining a new block and selectively reveal it when the public chain approaches the length of the private chain, making sure
that
the honest miners waste their computational power on their mined blocks.
Therefore, selfish miners can earn excessive rewards, which can be represented by the following MDP:
\begin{itemize}
    \item The state space $(l_a, l_h)$ is used to record the current state of the blockchain, where $l_a$ is the lengths of the private chain, and $l_h$ is the lengths of the public chain.
    \item The action space \{\textit{adopt, override, match, wait}\} represents the possible actions of the selfish miner in each state regarding mining and broadcasting blocks. Specifically, \textit{adopt} indicates abandoning the private chain and switching to mining on the longest chain, while \textit{override} refers to broadcasting two blocks from the private chain. Action \textit{match} represents broadcasting one block from the private chain, and \textit{wait} means not broadcasting and continuing mining.    
    \item The transition probability of the system is determined by parameter $\alpha$. For example, when a selfish miner chooses the action \textit{wait} in state $(l_a, l_h)$, the next state will be $(l_a + 1, l_h)$ with probability $\alpha$ (the selfish miner mines a new block) or $(l_a, l_h + 1)$ with probability $1 - \alpha$ (the honest miner mines a new block).
    \item The reward of the selfish (honest) miner is the number of blocks accepted by all parties that were mined by the selfish (honest) miner. The goal of the selfish miner is to maximize its proportion of the total reward.
\end{itemize}


\subsection{Security Threshold Analysis by MDP}
The security threshold analysis through the Markov Decision Process provides critical insights into blockchain protocol vulnerabilities by quantifying the minimum resource requirements for attackers to profit from strategic mining. 
This methodology systematically models state transitions, reward mechanisms, and strategic interactions, enabling rigorous evaluation of blockchain security boundaries.

\paragraph{Foundational Model on Selfish Mining.}
The seminal work \cite{eyal2014majority} established the theoretical framework for strategic mining by formalizing the \textit{selfish mining} strategy as an Markov Reward Process. 
Their analysis revealed Bitcoin's non-incentive compatibility: attackers with over 25\% hashing power could gain disproportionate rewards by selectively withholding blocks. 
Subsequent research expanded the strategic mining design space through novel attack vectors. 
\cite{nayak2016stubborn} proposed the \textit{stubborn mining} strategy, demonstrating a quarter profit increase over traditional selfish mining by persisting on private chains despite public chain dominance.
By considering this strategy space, the 25\% threshold was further refined by \cite{sapirshtein2016optimal}, who introduced an 
$\epsilon$-optimal algorithm to demonstrate that attackers could exploit vulnerabilities with only 23.2\% computational power, thereby lowering Bitcoin's security boundary.



\paragraph{Multi-Attacker Scenarios and Equilibrium Analysis.}
The emergence of multi-miner competition introduced new dimensions to security threshold analysis. 
\cite{liu2018strategy} developed the \textit{publish-n strategy}, reducing stale block rates by 26.3\% compared to selfish mining through deterministic block release patterns. 
Building on this,~\cite{marmolejo2019competing} introduced semi-selfish Mining, which imposed a two-block limit on private chains to lower the security threshold to 26.297\%. Their simulations also revealed an inverse relationship between attacker count and security thresholds. 
Complementing these studies, although solving the MDP game has been proven to be computationally complex ~\cite{deng2023complexity}, ~\cite{zhang2022insightful} designed a mining game model and proved that honest mining remains an equilibrium strategy when attackers' computational power stays below 33\%.


\paragraph{Ethereum Analysis and Methodological Advances.}
Ethereum's unique reward mechanism necessitated tailored MDP frameworks.
\cite{feng2019selfish} constructed a two-dimensional MDP incorporating uncle/nephew block rewards, identifying a 26\% security threshold. 
To address Ethereum's nonlinear reward structure,~\cite{zur2020efficient} proposed the Probability Termination Optimization (PTO) method, converting complex MDPs into solvable forms. 
This innovation reduced Ethereum's security threshold to 24.68\%, demonstrating the critical role of reward function design in protocol robustness.

%% file: Sections/New-3-RL.tex
\section{Reinforcement Learning Framework for Analyzing Strategic Mining}
\label{sec:RL}

\begin{table*}[ht]
    \centering
    \large
    \resizebox{\textwidth}{!}{%
        \begin{tabular}{|m{4cm}<{\centering} m{2cm}<{\centering} m{3cm}<{\centering} m{2.5cm}<{\centering} m{2cm}<{\centering} m{4.5cm}|}
            \hline
            \textbf{Literature} & 
            \textbf{Security Threshold} & 
            \textbf{Threshold Condition} & 
            \textbf{Method} & 
            \textbf{Consensus Type} & 
            \textbf{Description} \\ 
            \hline
            
            \cite{hou2019squirrl} & 
            0.25 & 
            \textbackslash & 
            Deep Q Network & 
            Bitcoin \& Ethereum PoW & 
            Recover optimal selfish mining strategy in Bitcoin and surpasses the existing selfish mining strategies in Ethereum. \\ 
            \hline
            
            \cite{bar2022werlman} & 
            {\begin{tabular}[b]{@{}c@{}}
                0.20 \\
                0.17 \\
                0.12 \\
            \end{tabular}} & 
            {\begin{tabular}[b]{@{}c@{}}
                3 minting rate halving \\
                4 minting rate halving \\
                5 minting rate halving \\
            \end{tabular}} & 
            Monte Carlo Tree Search \& DQN & 
            Bitcoin PoW & 
            Shows that transaction fee volatility will reduce blockchain security. \\ 
            \hline
            
            \cite{bar2023deep} & 
            {\begin{tabular}[b]{@{}c@{}}
                0.21 \\
                0.19 \\
            \end{tabular}} & 
            {\begin{tabular}[b]{@{}c@{}}
                0.5 petty compliant \\
                0.75 petty compliant \\
            \end{tabular}} & 
            Monte Carlo Tree Search \& DQN & 
            Bitcoin PoW & 
            Shows selfish miners can boost profits by bribing partially compliant miners. \\ 
            \hline
            
            \cite{sarenche2024deep} & 
            0.24198 & 
            \textbackslash & 
            DQN & 
            Longest-Chain Proof of Stake & 
            Uses DQN in LC-PoS protocols and shows that the security threshold for selfish proposing attacks is lower than that selfish mining. \\ 
            \hline
        \end{tabular}%
    }
    \caption{Research on security threshold of blockchain consensus protocols by reinforcement learning methods}
    \label{tab:sec-threshold}
\end{table*}

Through reinforcement learning (RL) technique,
an agent learns optimal strategies via its interactions with 
environment by selecting actions based on the current state and adjusting its behavior based on rewards or penalties. 
Similar
to many important applications~\cite{arulkumaran2017deep,nosratabadi2020data,zhao2022deep}, 
incentive mechanisms ensure security 
of blockchain protocols
by rewarding miners who follow the protocol.
Reinforcement learning method
helps user to simplify  the complex dynamics of blockchain protocols for better security analysis.

\subsection{Theoretical Foundations of RL}

The learning process in RL can be approached through two fundamental methods:

\begin{itemize}
    \item \textbf{Value-based Methods}: These methods indirectly select the optimal policy by calculating the value of each state. 
    For example, Q-learning \cite{watkins1989learning} is a value-based method that learns Q-values (state-action value functions) to evaluate the quality of actions.

    \item \textbf{Policy-based Methods}: These methods directly optimize the policy itself, rather than first learning a value function and then selecting a policy. Policy gradient methods are a common form of this approach.
\end{itemize}

The core objective of reinforcement learning is to maximize cumulative rewards, and this process is typically modeled as a Markov Decision Process~\cite{puterman2014markov}. 

Value-based methods evaluate the long-term return under a given state by a value function. The optimal policy is then derived by maximizing the value function:
\begin{align*}
    \pi^* = \arg \max_{\pi} V_{\pi}(s).
\end{align*}

Policy-based methods directly learn the policy $\Pi(a|s)$, which usually maps states to actions with the policy parameters $\theta$. 
The goal of policy optimization is to maximize the expected return:

\begin{align*}
    J(\theta) = \mathbb{E}_{\pi} \left[ \sum_{t=0}^{\infty} \gamma^t R_{t} \right].
\end{align*}

Many classical algorithms in reinforcement learning, such as Q-learning, Monte Carlo methods, and Temporal Difference (TD) learning, can be viewed as numerical estimation methods for the value function in an MDP.

\begin{itemize}
    \item \textbf{Q-learning}: It is a model-free method to find the optimal policy by updating Q values (state-action values). Although it does not require a model of the environment, it is similar to MDP’s state value function because both aim to optimize the expected future return. 
    \item \textbf{Monte Carlo Methods}: Monte Carlo methods estimate the value function by sampling the environment multiple times. Their computation process closely mirrors the value function calculation in an MDP, with the key difference being that they use actual returns rather than expected returns.
    \item \textbf{Temporal Difference (TD) Learning}: TD learning updates the estimate of the current state by incorporating the immediate reward and the value function of the next state at each step. It combines concepts from dynamic programming and the Bellman equation in an MDP.
\end{itemize}

\subsection{Strategic Mining Analysis through RL}



Building on the foundational concepts of RL in modeling strategic mining behaviors, this subsection delves into the application of RL-based frameworks to analyze and optimize mining strategies in blockchain protocols.
Specifically, we explore how RL has been employed to develop analytical frameworks, estimate advanced security thresholds, and devise novel attack strategies and countermeasures. 
These studies highlight the versatility of RL in capturing the complex dynamics of strategic mining, offering insights into both the vulnerabilities and potential mitigations within blockchain systems.
The key findings are summarized in Table~\ref{tab:sec-threshold}.

\paragraph{RL-Based Analytical Frameworks.}

Several studies have leveraged reinforcement learning to analyze and optimize strategic mining behaviors. SquirRL is one of these pioneering works. This analytical framework proposed in~\cite{hou2019squirrl} utilizes RL to analyze blockchain's incentive mechanisms. It defines agent capabilities and action spaces, creates a simulation environment, and incorporates elements such as agent extraction, RL algorithm selection, and reward function design. 
In turn, it successfully identified the optimal selfish mining attack in Bitcoin and discovered a novel attack on Ethereum’s Casper FFG protocol.
Another work~\cite{wang2021blockchain} explored the feasibility to dynamically learn optimal strategic mining approaches. Unlike conventional analytical models that require explicit parameter extraction, their approach employs RL to observe the blockchain network and consensus protocol, adapting mining strategies to time-varying network conditions without relying on prior knowledge of MDP parameters.



\paragraph{Advanced Security Threshold Estimation.}
Building on the application of RL, \cite{bar2022werlman} introduced WeRLman, an RL framework that incorporates "whales" (high-value transactions) and variance reduction techniques to more accurately estimate security thresholds. 
By using variance reduction to mitigate high sampling noise and optimizing strategies with Monte Carlo Tree Search, the framework determined Bitcoin’s security threshold to be approximately 25\% (which decreases over time) and Ethereum’s to be around 17\%. 
Expanding on this work, \cite{bar2023deep} extended WeRLman to explore the impact of small miners on blockchain security. 
They assumed a mix of compliant small miners and honest miners, and found that selfish miners could exploit weakened attack defenses, increasing their profits by over 10\%. 
In the Bitcoin scenario, when half of the miners were small and compliant, the security threshold dropped from 25\% to 21\%.

\paragraph{Alternative RL-Based Attacks and Countermeasures.}
Reinforcement learning has enabled novel attack strategies beyond selfish mining. \cite{yang2020ipbsm} introduced an intelligent bribery-based selfish mining attack, using RL to optimize strategies and outperform traditional models in profitability and equity thresholds. 
By modeling the environment as an MDP and leveraging RL-based decision-making, their approach surpassed traditional selfish mining models in terms of equity threshold and profitability.
Similarly, \cite{jeyasheela2024q} applied machine learning to predict miner rewards with high accuracy (MSE: 0.0032) and used Q-learning to simulate selfish mining behaviors.
Their $\epsilon$-greedy value iteration approach improved attacker profitability while offering insights into countermeasures. These studies highlight RL's dual role in advancing attacks and defenses in blockchain systems.

\paragraph{RL in Proof-of-Stake Blockchains.}
While strategic mining attacks have traditionally been associated with Proof-of-Work (PoW) blockchains, the longest chain rule is also employed in some Proof-of-Stake (PoS) protocols, where mining is replaced by proposer elections. 
This shift introduces a new attack vector, known as selfish proposing, which is analogous to selfish mining. \cite{sarenche2024deep} investigated selfish proposing attacks in longest chain PoS (LC-PoS) blockchains, analyzing how attackers exploit the "nothing-at-stake" problem and proposer predictability. 
The study found that the "nothing-at-stake" phenomenon slightly increases the proportion of blocks proposed by attackers, while the predictability of proposers significantly increases the proportion of attack blocks.
To analyze selfish proposing attacks in more complex scenarios, they also used deep Q-learning tools to approximate the optimal attack strategies under different stake shares.

%% file: Sections/New-4-Consensus_Strategy.tex
\section{Consensus Protocol Classification and Open Problems}
\label{sec:consensus}

Consensus protocols are fundamental to blockchain systems, ensuring that all participants agree on transaction validity and the blockchain’s state, thereby preventing unauthorized modifications and preserving the network’s integrity. 
These protocols can be classified on the basis of their chain selection rules, which include chain-based, vote-based, and DAG-based (parallel confirmation) approaches.
When analyzing strategic mining across different consensus protocols, a key step is constructing the environment, tailored to the specific incentive mechanism and underlying consensus algorithm. 
In this section, we will explore the similarities and differences in constructing environments for various consensus protocols and propose meaningful open questions for future research.

\subsection{Consensus Protocol Overview}

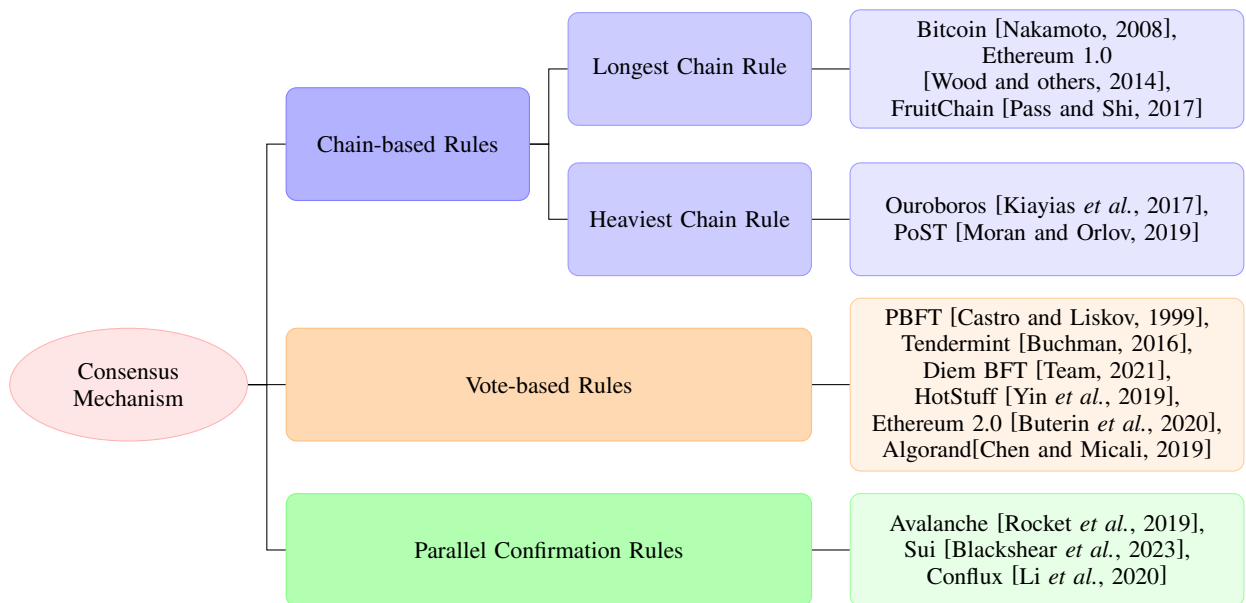
\begin{figure*}[ht]
\centering

\begin{tikzpicture}[  
    node distance=0.5cm, 
    every node/.style={align=center, font=\small, text centered, minimum height=1.5cm, rounded corners}, 
    every edge/.style={ultra thin, draw=black},  
]  

\node[ellipse, draw=red!30, fill=red!10, text width=2cm] (root) {Consensus Mechanism};  

\node[rectangle, draw=blue!50, fill=blue!30, text width=3cm, right=of root, yshift=3.2cm] (chain) {Chain-based Rules};  
\node[rectangle, draw=orange!50, fill=orange!30, text width=6.75cm, right=of root] (vote) {Vote-based Rules};  
\node[rectangle, draw=green!50, fill=green!30, text width=6.75cm, right=of root, yshift=-2.2cm] (parallel) {Parallel Confirmation Rules};  

\draw (root.east) -- ++(0.25,0) |- (chain.west);  
\draw (root.east) -- ++(0.25,0) |- (vote.west);  
\draw (root.east) -- ++(0.25,0) |- (parallel.west);  

\node[rectangle, draw=blue!40, fill=blue!20, text width=3cm, right=of chain, yshift=1cm] (longest) {Longest Chain Rule};  
\node[rectangle, draw=blue!40, fill=blue!20, text width=3cm, right=of chain, yshift=-1cm] (heaviest) {Heaviest Chain Rule};  

\draw (chain.east) -- ++(0.25,0) |- (longest.west);  
\draw (chain.east) -- ++(0.25,0) |- (heaviest.west);  

\node[rectangle, draw=blue!40, fill=blue!10, text width=5cm, right=of longest] (bitcoin) {Bitcoin \cite{nakamoto2008bitcoin},\\ Ethereum 1.0 \cite{wood2014ethereum},\\ FruitChain \cite{pass2017fruitchains}};  
\node[rectangle, draw=blue!40, fill=blue!10, text width=5cm, right=of heaviest] (ouroboros) {Ouroboros \cite{kiayias2017ouroboros}, PoST \cite{moran2019simple}};  
\node[rectangle, draw=orange!40, fill=orange!10, text width=5cm, right=of vote] (pbft) {PBFT \cite{castro1999practical}, \\Tendermint \cite{buchman2016tendermint}, \\Diem BFT \cite{team2021diembft},\\ HotStuff \cite{yin2019hotstuff},\\ Ethereum 2.0 \cite{buterin2020combining},\\Algorand\cite{chen2019algorand}};  
\node[rectangle, draw=green!40, fill=green!10, text width=5cm, right=of parallel] (avalanche) {Avalanche \cite{rocket2019scalable}, \\ 
Sui \cite{blackshear2023sui}, \\
Conflux \cite{li2020decentralized}};  

\draw (longest.east) -- ++(0.25,0) |- (bitcoin.west);  
\draw (heaviest.east) -- ++(0.25,0) |- (ouroboros.west);  
\draw (vote.east) -- ++(0.25,0) |- (pbft.west);  
\draw (parallel.east) -- ++(0.25,0) |- (avalanche.west);  
\end{tikzpicture}
\caption{Blockchain Consensus Protocol Classification}
\end{figure*}

Blockchain consensus mechanisms can be classified based on their chain selection rules, including chain-based rules, vote-based rules, DAG-based rules.

\paragraph{Chain-based Consensus Rules.} 
Chain-based consensus rules rely on a linear blockchain structure, where blocks are linked in a chain, and the main chain is determined by the accumulated work or weight. 
The Longest Chain Rule, used in Proof-of-Work (PoW) systems like Bitcoin \cite{nakamoto2008bitcoin} and Ethereum 1.0 \cite{wood2014ethereum}, selects the chain with the most computational work. FruitChain \cite{pass2017fruitchains} extends the Longest Chain Rule by incorporating a dual-reward system, where miners earn rewards not only for block creation but also for broadcasting ``fruit'' structures, thus enhancing system security. 
Similarly, the Heaviest Chain Rule selects the chain with the greatest accumulated weight, typically based on stake, as seen in systems like Ouroboros \cite{kiayias2017ouroboros}. Additionally, Proof of Space and Time (PoST) \cite{moran2019simple} relies on the amount of storage and time invested to secure the chain, using a similar structure where the "heaviest" chain is the one that accumulates the most storage and time.

\paragraph{Vote-based Consensus Rules.} 
These consensus protocols involve nodes casting votes on the validity of transactions or blocks. 
In vote-based consensus systems, such as Practical Byzantine Fault Tolerance (PBFT) \cite{castro1999practical}, Tendermint \cite{buchman2016tendermint}, Algorand \cite{chen2019algorand} and HotStuff \cite{yin2019hotstuff}, a multi-phase voting process ensures that once a block is committed, it cannot be reverted, thus preventing forks. 
These protocols offer strong finality guarantees, with blocks being immediately finalized once they receive a sufficient number of votes.

\paragraph{Parallel Confirmation Rules.} Parallel confirmation rules deviate from traditional chain-based structures by allowing multiple branches to exist in parallel. 
In protocols like Avalanche \cite{rocket2019scalable}, Conflux \cite{li2020decentralized}, and Sui \cite{blackshear2023sui}, consensus is achieved probabilistically, with transactions validated concurrently across different branches. 
Since there is no main chain, consensus is reached without relying on a single-chain structure. 

It is also important to note that some consensus mechanisms may exhibit both primary and secondary attributes, blending features from different categories. 
For example, Ethereum 2.0 \cite{buterin2020combining} primarily relies on vote-based consensus (LMD-GHOST), while also incorporating the heaviest chain rule (via stake-weighted mechanisms) as a secondary feature.

\subsection{Design Components Across Different Consensus Protocols}

In this section, we outline the similarities and differences in environment construction for various consensus protocols, highlighting key design components such as \textit{state space}, \textit{action space}, and \textit{reward design}.

\paragraph{State Space.}
The state space must encode three critical components: (1) Action availability: features representing permissible actions in the current state, such as the fork status in Bitcoin (to track competing chains) \cite{sapirshtein2016optimal} or the match flag indicating active participation in protocols like LC-PoS \cite{sarenche2024deep}.
(2) Reward computation: features enabling reward calculation based on the canonical chain or subgraph. For example, in Bitcoin, this involves tracking the lengths of competing chains ($l_a, l_h$) to resolve forks \cite{nakamoto2008bitcoin}. For protocols with non-linear reward mechanisms (e.g., FruitChain’s "fruits" \cite{pass2017fruitchains,zhang2019lay} or Ethereum 2.0’s attestations \cite{zhang2024max}), additional metrics are required to compute relative rewards.
(3) State transition: The system state evolves through block generation, a discrete-time process with probability determined by mining power in PoW or stake in PoS. Transitions follow the consensus protocol’s stochastic rules and network assumptions, including idealized instant block propagation. During ties, honest nodes adopt adversarial blocks with a probability (rushing factor), modeling latency exploitation.


\paragraph{Action Space.}
The design of the action space is closely tied to the adversarial model. 
For each consensus mechanism, different types of adversaries can be defined. 
For example, in the Bitcoin protocol, the action space is defined as {\textit{adopt, override, wait, match}}. 
Other attack strategies may involve actions outside this predefined space, such as the consideration of petty compliant miners in \cite{bar2023deep}. 
In the selfish proposing attack targeting the LC-PoS protocol \cite{sarenche2024deep}, the action space includes sub-actions to capture the `jump' strategy, allowing a selfish proposer to alter the parent block due to the ``nothing-at-stake'' property. 
In Ethereum 2.0, the action space may exclude the `match' action, as the rule for determining the canonical chain is based on the heaviest chain rather than the longest chain. 
This distinction removes the need to consider network propagation when competing chains have equal block lengths.

\paragraph{Reward Design.}
In reward design, previous approaches have primarily focused on relative rewards, defined as the attacker's rewards as a fraction of the total network rewards. 
These rewards are typically established on a per-unit basis, such as for blocks, as seen in earlier works. 
However, rewards can also be more intricately characterized, including transaction rewards \cite{bar2022werlman} and attestation rewards in Ethereum 2.0 \cite{zhang2024max}. 
The goal of the analysis is to identify a strategy $\pi: S \to \Delta(A)$ that maximizes the expected reward $R(s,a)$. 


\subsection{Open Problems}
The evolution of blockchain technology and the digital economy has led to a significant shift from traditional longest-chain consensus mechanisms to alternative models. 
However, the economic security of these systems hinges on resolving critical open problems that remain inadequately addressed.
Moreover, the growing sophistication of miners in employing strategic mining techniques has introduced the risk of multi-agent strategic behaviors. These behaviors could destabilize the ecosystem, leading to economic inefficiencies or even systemic failures.
This section identifies important open problems in using RL for blockchain security analysis. 
Each problem highlights the need for advanced modeling techniques to address the complexities of modern blockchain systems.

\paragraph{Open problem 1: How can extend strategic mining attack analysis to non-longest-chain consensus protocols?}

Existing research on strategic mining attacks has predominantly centered on longest-chain consensus protocols~\cite{eyal2014majority,sapirshtein2016optimal,sarenche2024deep}. 
To generalize this framework to alternative consensus mechanisms, three directions emerge:
\begin{itemize}
    \item \textbf{Weight-Based Protocols:} The state space must explicitly model weight accumulation dynamics. This requires incorporating weight-related parameters into the strategy space while preserving backward compatibility with existing analysis methods for longest-chain systems.
    \item \textbf{Parallel Proof-of-Work Protocols:}  The state space can be generalized to include additional features, such as topological order and uncle block rewards. These additions introduce multi-dimensional optimization challenges absent in linear chain protocols.
    \item \textbf{Vote-Based Consensus:} Participants attempt to minimize the costs associated with validating blocks and sending votes, which can result in coordination failures that undermine the validity of consensus protocols \cite{amoussou2020governing}. Furthermore, attackers can execute censorship attacks \cite{srivastava2024towards} by strategically excluding specific information from being incorporated into the final consensus.
\end{itemize}
Multiple tricks such as imposing artificial limits within the adversarial model \cite{sapirshtein2016optimal,zur2020efficient,hou2019squirrl} can help maintain a manageable state space size.
The application of the analysis framework to adversarial models targeting the aforementioned attacks still requires further exploration.



\paragraph{Open problem 2: How to develop more realistic MDP models for blockchain security?}

Current RL models for blockchain security rely on simplified assumptions, such as synchronized networks and fixed miner strategies \cite{zhang2019lay,sarenche2024deep}, to reduce complexity. However, in real-world environments, network conditions, miner strategies, and blockchain dynamics are highly unpredictable. For example, attackers may exploit network latency to conduct undetectable attacks, limiting the applicability of existing models \cite{bahrani2023undetectable}.

Future research should focus on relaxing these assumptions by incorporating dynamic, time-varying environments. RL models can account for unpredictable network delays, adaptive miner strategies, and changing blockchain dynamics. 
These uncertain conditions can be handle by using advanced RL algorithms, while ensuring robust security analysis under evolving threats.

\paragraph{Open problem 3: How can strategic mining be analyzed in multi-agent environments using RL?}
A key challenge is applying RL in multi-agent environments, where multiple miners or validators interact strategically to maximize rewards. In these environments, agents may compete, complicating the analysis of selfish mining attacks. 

\cite{marmolejo2019competing} use a Markov chain model to analyze multi-agent mining dynamics by simplifying the state space of selfish miners. This analysis restrict to semi-selfish mining, where miners only maintain private chains of length at most two. 
The Partially Observed Markov Game (POMG) extends MDP to multi-agent environments with partial information, allowing it to model strategic behaviors like selfish mining, such as SquirRL \cite{hou2019squirrl}. 
However, POMG has limitations, including assumptions about partial observability and challenges with scalability as the number of miners increases. Future research should focus on improving its scalability and refining agent interaction models to better capture the complexities and unpredictability of real-world blockchain dynamics.